\documentclass[letterpaper]{article} 
\usepackage{aaai2026}  
\usepackage{times}  
\usepackage{helvet}  
\usepackage{courier}  
\usepackage[hyphens]{url}  
\usepackage{graphicx} 
\usepackage{natbib}  
\usepackage{caption} 
\usepackage{algorithm}
\usepackage{algorithmic}
\usepackage{amsmath}
\usepackage{array}  
\usepackage[table]{xcolor}
\usepackage{multirow}
\usepackage{booktabs}
\usepackage{amssymb}
\usepackage{tabularx}
\usepackage{newfloat}
\usepackage{listings}
\DeclareCaptionStyle{ruled}{labelfont=normalfont,labelsep=colon,strut=off} 
\floatstyle{ruled}
\newfloat{listing}{tb}{lst}{}
\floatname{listing}{Listing}
\title{Decoding with Structured Awareness: Integrating Directional, Frequency-Spatial, and Structural Attention for Medical Image Segmentation}
\author{
    Fan Zhang\textsuperscript{\rm 1},
    Zhiwei Gu\textsuperscript{\rm 1},
    Hua Wang\textsuperscript{\rm 2}\thanks{Corresponding author.}
}
\affiliations{
    \textsuperscript{\rm 1}School of Computer Science and Technology, Shandong Technology and Business University, Yantai 264005, China \\
    \textsuperscript{\rm 2}School of Computer and Artificial Intelligence, Ludong University, Yantai 264025, China \\
    zhangfan@sdtbu.edu.cn, guzhiwei409@gmail.com, hwa229@163.com
}

\begin{document}

\maketitle

\begin{abstract}
To address the limitations of Transformer decoders in capturing edge details, recognizing local textures and modeling spatial continuity, this paper proposes a novel decoder framework specifically designed for medical image segmentation, comprising three core modules. First, the Adaptive Cross-Fusion Attention (ACFA) module integrates channel feature enhancement with spatial attention mechanisms and introduces learnable guidance in three directions (planar, horizontal, and vertical) to enhance responsiveness to key regions and structural orientations. Second, the Triple Feature Fusion Attention (TFFA) module fuses features from Spatial, Fourier and Wavelet domains, achieving joint frequency-spatial representation that strengthens global dependency and structural modeling while preserving local information such as edges and textures, making it particularly effective in complex and blurred boundary scenarios. Finally, the Structural-aware Multi-scale Masking Module (SMMM) optimizes the skip connections between encoder and decoder by leveraging multi-scale context and structural saliency filtering, effectively reducing feature redundancy and improving semantic interaction quality. Working synergistically, these modules not only address the shortcomings of traditional decoders but also significantly enhance performance in high-precision tasks such as tumor segmentation and organ boundary extraction, improving both segmentation accuracy and model generalization. Experimental results demonstrate that this framework provides an efficient and practical solution for medical image segmentation.
\end{abstract}

\section{Introduction}

Medical image segmentation plays a pivotal role in intelligent healthcare and clinical applications, aiming to accurately delineate organs, tumors, or lesions from complex medical images, thereby providing clinicians with structured and intuitive reference information. This not only significantly improves the accuracy of diagnosis and lesion assessment but also offers crucial support for surgical planning, radiotherapy dose design, and treatment monitoring. With the rapid growth of medical imaging data, automated segmentation methods have become indispensable for reducing clinicians’ workload, enhancing diagnostic efficiency, and ensuring result consistency. Deep learning, as a driving force of artificial intelligence, has greatly advanced the application of Convolutional neural networks (CNNs) in image segmentation. Early models such as LeNet \cite{LeNet} and AlexNet \cite{AlexNet} demonstrated the advantages of deep convolutional architectures in feature extraction. Subsequent models like VGG \cite{VGG} and ResNet \cite{ResNet} enhanced feature representation and training stability through residual and multi-path designs. Building on this foundation, U-Net \cite{unet} proposed an encoder-decoder architecture that captures global semantic information through downsampling and recovers spatial details via skip connections that fuse multi-level features. However, traditional skip connections often rely on simple addition operations, which may lead to the loss of spatial details and the inclusion of redundant information, making it difficult to balance global and local features. In recent years, Vision Transformers have shown great potential in medical image segmentation due to their ability to capture long-range dependencies through self-attention mechanisms. Representative methods such as Swin-UNet \cite{Swin-unet}, PVT \cite{PVT}, MaxViT \cite{Maxvit}, MERIT \cite{MERIT}, and ConvFormer \cite{Convformer} have improved overall segmentation performance but still struggle with modeling fine-grained textures and edge details, and the development of large models has also provided insights for our research \cite{MedSAM}. Furthermore, recent developments in composed image retrieval, including ENCODER \cite{ENCODER}, FineCIR \cite{FineCIR}, OFFSET \cite{OFFSET}, HUD \cite{HUD}, PAIR \cite{PAIR}, and MEDIAN \cite{MEDIAN}, reveal that integrating structured reasoning with uncertainty modeling and hierarchical feature aggregation can significantly improve representation robustness and semantic interpretability, inspiring advances in medical image segmentation. Therefore, inspired by the aforementioned advances, we propose a novel decoder framework to address these challenges by preserving global perception while enhancing the representation of edges and structural details. This decoder consists of three key modules:
\begin{itemize}
    \item ACFA: A direction-aware module that strengthens structural orientation and spatial consistency.
    \item TFFA: A tri-branch fusion module integrating spatial, Fourier, and wavelet representations to balance global and local features.
    \item SMMM: A multi-scale skip-fusion module that suppresses redundancy, refines boundaries, and improves feature alignment for accurate and detailed segmentation.
\end{itemize}

\section{Related Works}
\subsection{Vision Encoders}
CNNs are widely used as encoders in medical image segmentation for their efficiency in extracting multi-level spatial and semantic features through local convolutions, weight sharing, and mature architectures. VGGNet \cite{VGG} showed that stacking small kernels with increased depth improves performance, while GoogLeNet \cite{GoogLeNet} employs Inception modules and global average pooling to achieve efficient feature extraction with fewer parameters. ResNet \cite{ResNet} introduced residual connections to overcome degradation in deep networks, enabling deeper models with high efficiency. However, CNNs’ fixed receptive fields limit their ability to capture long-range dependencies and global context, which is critical for complex shapes or blurred boundaries. Vision Transformers (ViTs) address this via self-attention, with variants like Swin Transformer \cite{swinTransformer} and PVT \cite{PVT} enhancing global modeling. PVTv2 \cite{Pvtv2} further combines convolutions for local features, overlapping patch embedding, and linear attention with mean pooling. Despite improvements in global representation, these methods still struggle with modeling short-range dependencies effectively.

\subsection{Medical Image Segmentation}
\begin{figure*}
    \centering
    \includegraphics[height=0.6\linewidth]{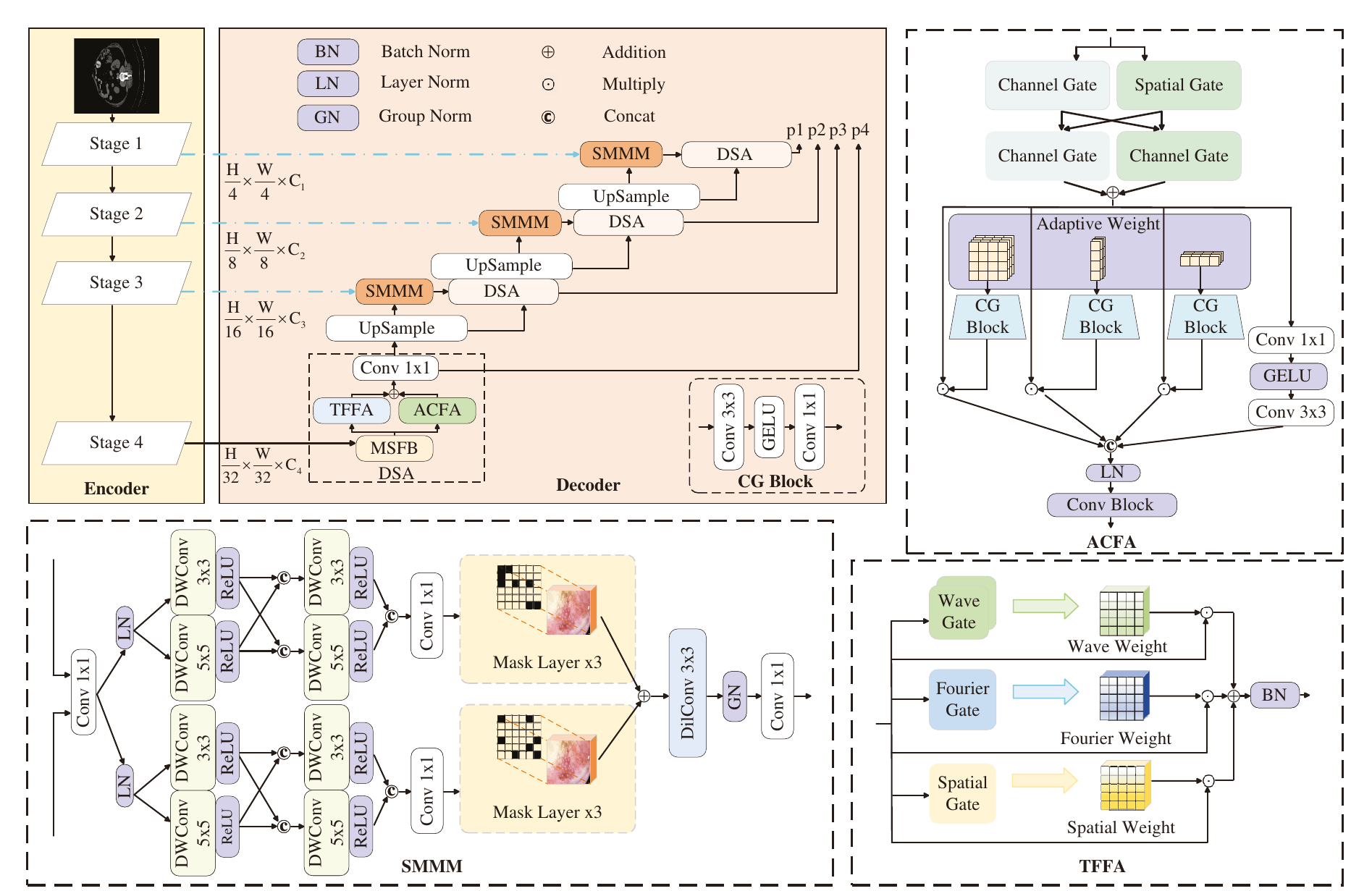}
    \caption{Proposed network architecture of the model.}
    \label{fig:main}
\end{figure*}

Medical image segmentation, a key task in medical image analysis, benefits significantly from deep learning’s capability for powerful feature extraction and end-to-end modeling. Classic CNN-based frameworks such as U-Net \cite{unet}, V-Net \cite{V-net}, SegNet \cite{Segnet}, and FCN \cite{fcn} have laid the foundation for automated delineation of organs and lesions. U-Net’s encoder–decoder architecture with skip connections enables multi-scale fusion, while V-Net extends this idea to 3D data, and SegNet and FCN improve spatial preservation and end-to-end prediction, respectively. Recent advances further enhance representational capacity: AD-LA Former \cite{AD-LA} combines dynamic convolutions and positional attention for complex structures, whereas EMCAD \cite{Emcad} introduces a multi-scale attention decoder that strengthens salient region modeling through channel, spatial, and grouped gating attention. Transformer-based models such as TransUNet \cite{TransUNet}, Swin-UNet \cite{Swin-unet}, and CSWin-UNet \cite{CSWin-UNet} extend long-range contextual modeling via self-attention mechanisms, while MISSFormer \cite{missformer} and LeViT-UNet \cite{Levit-unet} improve efficiency through multi-scale and lightweight design. Hybrid architectures that combine CNNs and Transformers—such as Hiformer \cite{hiformer} and TBConvL-Net \cite{TBConvL-Net}—leverage the strengths of both paradigms to balance locality, global dependency, and computational cost. Beyond conventional segmentation models, progress in efficient learning \cite{swift-sampler}, federated harmonization \cite{confusion-resistant-FL}, cross-modal contrastive learning \cite{C3-OWD}, and diffusion-based self-supervision \cite{diffusion-robotic} has inspired new directions for robust feature representation. Moreover, advancements in normalized 3D scene modeling \cite{NDC-scene} and multimodal reasoning \cite{Simignore} provide insights into efficient information fusion and reasoning that could further benefit medical image segmentation.

\section{Methods}
Our goal is to develop a novel medical image segmentation decoder that can effectively preserve the integrity of both local and global information while placing greater emphasis on key regional features. The following sections provide a detailed description of the decoder’s structural design and functionality, as shown in the Figure \ref{fig:main}, which consists of three core modules:

\subsection{Adaptive Cross-Fusion Attention}
To enhance the model's responsiveness to key regions and its ability to model structural directions, we design an ACFA module with directional awareness. This module integrates channel enhancement and spatial modeling mechanisms while introducing learnable guidance along three directions (planar, vertical, and horizontal), enabling fine-grained feature enhancement across different scales and orientations. Specifically, for an input feature map $X\in {{\mathbb{R}}^{B\times C\times H\times W}}$, channel and spatial gating are first applied to extract features:
\begin{align}
  & \widehat{X}_{l-1}^{CG}=X\odot \delta \left( C{{G}_{avg}}\left( X \right)+C{{G}_{\max }}\left( X \right) \right) 
\end{align}
\begin{align}
 & \widehat{X}_{l-1}^{SG}=X\odot \delta \left( f_{7\times 7}^{Conv}\left( SG\left( X \right) \right) \right) 
\end{align}
Here, $C{{G}_{avg}}\left( {} \right)$ and $C{{G}_{\max }}\left( {} \right)$ represent average channel gating and maximum channel gating, respectively. $SG\left( {} \right)$ denotes spatial feature gating, $f_{7\times 7}^{Conv}\left( {} \right)$ refers to a convolution operation with a $7\times 7$ kernel, $\delta \left( {} \right)$ is the Sigmoid activation function, and $\odot $ indicates matrix multiplication. Subsequently, to obtain deeper-level features, the feature map is divided along the channel dimension into four subsets $\widehat{X}_{l-1}^{S{{G}_{1}}}, \widehat{X}_{l-1}^{S{{G}_{2}}}, \widehat{X}_{l-1}^{S{{G}_{3}}}, \widehat{X}_{l-1}^{S{{G}_{4}}}$, each of which is combined with learnable parameters corresponding to specific directions for fine-grained modulation. Depthwise separable convolutions are then applied to extract critical responses for each direction. Specifically, three branches introduce learnable weight parameters with directional guidance:
\begin{align}
  & Tenso{{r}^{HW}}=f_{\left[ 0,1 \right]}^{Uniform}\left( Param\left[ 1,\frac{C}{4},H,W \right] \right) 
  \end{align}
\begin{align}
 & Tenso{{r}^{H}}=f_{\left[ 0,1 \right]}^{Uniform}\left( Param\left[ 1,\frac{C}{4},H,1 \right] \right) 
 \end{align}
\begin{align}
 & Tenso{{r}^{W}}=f_{\left[ 0,1 \right]}^{Uniform}\left( Param\left[ 1,\frac{C}{4},1,W \right] \right)  
\end{align}
Here, $f_{\left[ 0,1 \right]}^{Uniform}\left( {} \right)$ denotes initialization using a Uniform function, which sets the tensor values to random numbers within the range $[0,1]$. These tensors are then processed through depthwise separable convolutions to further model the structural distribution patterns in different directions. The parameters are optimized in an end-to-end manner, enabling the model to automatically learn the most suitable directional attention patterns for the data distribution during training. Specifically:
\begin{align}
  & \widehat{X}_{l-1}^{HW}=f_{3\times 3}^{DWConv}\left( \vartheta \left( f_{1\times 1}^{Conv}\left( Tenso{{r}^{HW}} \right) \right) \right) \\ 
 & \widehat{X}_{l-1}^{H}=f_{3}^{DWConv1d}\left( \vartheta \left( f_{1}^{Conv1d}\left( Tenso{{r}^{H}} \right) \right) \right) \\ 
 & \widehat{X}_{l-1}^{W}=f_{3}^{DWConv1d}\left( \vartheta \left( f_{1}^{Conv1d}\left( Tenso{{r}^{W}} \right) \right) \right)  
\end{align}
Here, $f_{3}^{DWConv1d}\left( {} \right)$ and $f_{1}^{Conv1d}\left( {} \right)$ represent 1D depthwise separable convolution with a kernel size of 3 and 1D convolution with a kernel size of 1, respectively, while $\vartheta \left( {} \right)$ denotes the GELU activation function. To complement any detail information that might be missed in other directions, the fourth branch employs a set of standard convolution operations to capture more generalized contextual information, specifically:
\begin{align}
& \widehat{X}_{l-1}^{4}=f_{1\times 1}^{Conv}\left( \vartheta \left( f_{3\times 3}^{DWConv}\left( \widehat{X}_{l-1}^{S{{G}_{4}}} \right) \right) \right)
\end{align}
Finally, the features from the three directional branches and the fourth branch are concatenated, followed by LayerNorm and convolutional fusion to obtain the final direction-aware enhanced output.

\begin{table*}[ht]
\centering
\begin{tabular}{@{}l|ll|llllllll@{}}
\toprule \hline
\multirow{2}{*}{Methods} & \multicolumn{2}{c|}{Average}  & \multirow{2}{*}{Spl} & \multirow{2}{*}{RKid} & \multirow{2}{*}{LKid} & \multirow{2}{*}{Gal} & \multirow{2}{*}{Liv} & \multirow{2}{*}{Sto} & \multirow{2}{*}{Aor} & \multirow{2}{*}{Pan} \\ \cline{2-3}
                       & DSC ↑       & HD95  ↓                  
                       &                      &                       &                      
                       &                      &                      &                      
                       &                      &                     \\  \hline \hline
TransUNet    & 77.49  & 31.69   & 85.08 & 77.02 & 81.87 & 63.16 & 94.08 & 75.62 & 87.23 & 55.86 \\
Swin-UNet     & 79.13  & 21.55   & 90.66 & 79.61 & 83.28 & 66.53 & 94.29 & 76.60  & 85.47 & 56.58 \\
LeViT-UNet & 78.53  & 78.53   & 88.86 & 80.25 & 84.61 & 62.23 & 93.11 & 72.76 & 87.33 & 59.07 \\
MISSFormer    & 81.96  & 18.20    & 91.92 & 82.00    & 85.21 & 68.65 & 94.41 & 80.81 & 86.99 & 65.67 \\
ScaleFormer   & 82.86  & 16.81   & 89.40  & 83.31 &86.36 & 74.97 &95.12 & 80.14 &88.73 & 64.85 \\
HiFormer-B    & 80.39  & {\textbf{14.70} }    & 90.99 & 79.77 & 85.23 & 65.23 & 94.61 & 81.08 & 86.21 & 59.52 \\
DAEFormer    & 82.63  &16.39   &91.82 &82.39 & 87.66 & 71.65 & 95.08 & 80.77 & 87.84 & 63.93 \\
PVT-CASCADE   & 81.06  & 20.23   & 90.10  & 80.37 & 82.23 & 70.59 & 94.08 &83.69 & 83.01 & 64.43 \\
LKA           & 82.77  & 17.42   & 91.45 & 81.93 & 84.93 & 71.05 & 94.87 & 83.71 & 87.48 &66.76 \\
EMCAD   & 83.63  & 15.68   & 92.17    &84.10 & 88.08 & 68.87 & {\textbf{95.26} } & {\textbf{83.92} }  & 88.14 & 68.51 \\
AD-LA Former       & 83.48  &21.31   & 88.72 & 70.82 & 86.50 & {\textbf{83.30} } & 95.17 & 67.28 & {\textbf{91.34} } & {\textbf{84.69} } \\
CSWin-UNet    & 81.12  & 18.86   & 89.05 & 78.53 & 83.51 & 67.85 & 95.23 & 81.74 & 87.13 & 65.94 \\ \hline
 \textbf{Ours}           & {\textbf{83.92} }  & 18.91   & {\textbf{92.46}} & {\textbf{86.47} }  & {\textbf{89.26} } & 67.51 & 94.72 & 83.33 & 87.63 & 69.95\\  \hline \hline
\end{tabular}
\caption{The comparison results of the model with previous methods on the Synapse dataset. Bold indicates the best result, dsc is presented for abdominal organs spleen (Spl), right kidney (Rkid), left kidney (Lkid), gallbladder (Gal), liver (Liv), stomach (Sto), aorta (Aor), and pancreas (Pan). the same below.
}
\label{tab:SY}
\end{table*}

\begin{table}[ht]
\centering
\small
\begin{tabular}{@{}l|llll@{}}
\toprule \hline
      Methods    & DSC   & SE    & SP      & ACC  \\  \hline \hline
                 TransUNet       & 81.23 & 82.63 & 95.77   & 92.07 \\
                 Swin-Unet       & 88.15 & 83.64 & {\textbf{98.69} }   & 95.82 \\
                 LKA             &90.99 & 90.55 & 98.49   & 96.98 \\
                 Hiformer-B          & 90.93 & 88.67 &98.57   &96.69 \\
                 EGE-Unet        & 89.28 & 88.39 & 98.11   & 96.48 \\
                 UltraLightVM-UNet & 90.91 & 90.53 & 97.90    & 96.46 \\
                 EMCAD       & 90.06 & {\textbf{93.70} }  & 96.81   & 96.55 \\
                 AD-LA Former   & 87.68 & 93.18  & 98.65   & 97.03 \\
                 TBConvL-Net   & 90.89 & 91.19 & 97.61   & 96.07 \\  \hline
 \textbf{Ours}          & {\textbf{91.40} } &92.75 & 97.78   & {\textbf{97.26} } \\  \hline \hline 
\end{tabular}

\caption{Comparison experiments on the ISIC 2017 dataset.}
\label{tab:17}
\end{table}

\begin{table}[ht]
\centering
\small
\begin{tabular}{@{}l|llll@{}}
\toprule  \hline
                  Methods    & DSC   & SE    & SP      & ACC  \\  \hline \hline
                 TransUNet      & 84.99 & 85.78   & 96.53 & 94.52 \\
                 Swin-Unet        & 86.03 & 79.98   & 98.53 & 94.45 \\
                 LKA               & 88.88 & 84.75   &98.34 & 95.34 \\
                 Hiformer-B        & 88.10  & 84.61   & 97.74 & 94.85 \\
                 EGE-Unet       & 89.04 &90.44   & 95.91 & 94.58 \\
                 VM-UNetV2        &89.73 & 88.64   & 97.13 & 95.06 \\
                 UltraLightVM-UNet & 89.40  & 86.80    & 97.81 & 95.58 \\
                 EMCAD        & 89.73 & 92.43  & 96.75 &  96.17 \\
                 DMSA-Unet         & 90.63 & {\textbf{97.14} }  & 97.14 & 96.16 \\
                 AD-LA Former     & 85.91 & 91.24  & {\textbf{99.04} } & 96.46 \\
                 \hline
 \textbf{Ours}            & {\textbf{90.71} }  & 93.34 & 96.18 & {\textbf{96.62} } \\  \hline \hline 
\end{tabular}
\caption{Comparison experiments on the ISIC 2018 dataset.} 
\label{tab:18}
\end{table}

\subsection{Triple Feature Fusion Attention}
Traditional spatial convolution structures rely on local convolution kernels with fixed receptive fields, which can capture local texture information but are limited in modeling long-range dependencies, global structural relationships, and cross-scale semantic interactions. In particular, during feature fusion, simple spatial stacking or averaging operations, although preserving spatial distribution information, tend to cause semantic smoothing and redundancy, which can obscure key edges or subtle regions and weaken the model’s sensitivity to details and boundaries. To address these limitations, we propose the TFFA module, which consists of three branches: a wavelet branch, a Fourier branch, and a spatial feature branch. In the wavelet branch, the module employs DoG and Mexican Hat wavelet functions for local spatio-frequency analysis. Both of these wavelet functions are classic filtering or edge-detection operators with strong spatial localization and frequency analysis capabilities. Compared to traditional wavelets such as Haar or Daubechies, they provide superior edge and texture representation. For the input tensor $X\in {{\mathbb{R}}^{B\times C\times H\times W}}$, the wavelet transform is applied as follows:
\begin{align}
W\left( a,b \right)=\int_{-\infty }^{\infty }{x\left( t \right){{\psi }_{a,b}}\left( t \right)}dt
\end{align}
where ${{\psi }_{a,b}}\left( {} \right)$ denotes the mother wavelet function, instantiated with different wavelet types. DoG (Difference of Gaussians) is defined as the difference between two Gaussian-blurred images of different scales:
\begin{align}
Dog\left( x,y \right)={{G}_{{{\sigma }_{1}}}}\left( x,y \right)-{{G}_{{{\sigma }_{2}}}}\left( x,y \right),{{\sigma }_{1}}<{{\sigma }_{2}}
\end{align}
It emphasizes image details within specific frequency bands, serving as a band-pass filter approximating LoG. By adjusting the Gaussian kernel parameter $\sigma $, specific texture scales can be enhanced. Medical images often have low contrast and blurred textures; DoG can highlight areas with significant grayscale changes, enhancing edge and contour perception. Its formula is:
\begin{align}
\psi _{a,b}^{Dog}\left( x \right)=-\frac{1}{\sqrt{a}}\left( \frac{x-b}{a} \right){{e}^{-\frac{{{\left( \frac{x-b}{a} \right)}^{2}}}{2}}}
\end{align}
Here, a is the scale parameter and b is the shift parameter, both of which are learnable, enabling dynamic fusion of multi-wavelet features. Mexican Hat, in contrast, detects edge zero-crossings by taking the second derivative after smoothing, while effectively suppressing noise interference:
\begin{align}
\psi _{a,b}^{MH}\left( x \right)=\frac{2}{\sqrt{3a}{{\pi }^{\frac{1}{4}}}}\left( 1-{{\left( \frac{x-b}{a} \right)}^{2}} \right){{e}^{-\frac{{{\left( \frac{x-b}{a} \right)}^{2}}}{2}}}
\end{align}
The wavelet transform provides localized multi-scale information, whereas the Fourier transform is inherently global, capturing overall structures and modeling long-range dependencies, thus compensating for the limited perception of large-scale structures in convolutional models. It transforms the image from the spatial domain to the frequency domain, representing periodic structures with sine and cosine waves. The frequency-domain features are modulated by learnable weight matrices, where high-frequency components correspond to edges and textures, and low-frequency components represent contours and backgrounds. Its formula is:
\begin{align}
F\left( u,v \right)=\iint{f\left( x,y \right){{e}^{-j2\pi \left( ux+vy \right)}}dxdy}
\end{align}
In the spatial branch, pointwise convolution is applied to extract spatial features. Unlike simple channel stacking or averaging, TFFA introduces an attention gating mechanism after the three branches, assigning dynamic weights to spatial, frequency, and wavelet outputs to achieve adaptive fusion. This mechanism avoids over-smoothing caused by traditional fusion strategies and enhances the model's response to salient regions in complex scenarios. The final fused result is processed with batch normalization and activation functions to improve feature stability and nonlinear modeling capability, thereby achieving superior performance in tasks involving complex boundaries and fine-grained segmentation.

\subsection{Structural-Aware Multi-Scale Masking Module}
In medical image segmentation, skip connections are essential for linking multi-scale features between the encoder and decoder and mitigating spatial detail loss during upsampling. However, traditional skip connections often use simple feature addition, lacking selective fusion and leading to redundant or irrelevant information. To overcome this limitation, we propose SMMM, a structure-aware multi-scale fusion strategy that strengthens feature representation and discrimination. Specifically, for encoder features $X\in {{\mathbb{R}}^{B\times C\times H\times W}}$ and decoder features $Y\in {{\mathbb{R}}^{B\times C\times H\times W}}$ of the same shape, both are first processed by parallel pointwise convolutions to activate spatial cues, followed by multi-scale perception modules. Each module integrates depthwise separable convolutions with kernel sizes ($3\times 3$, $5\times 5$), combined with a two-stage channel split and ReLU activation, effectively enlarging the receptive field and improving the ability to capture complex structural boundaries. The detailed formulation is as follows:

\begin{align}
  & {{\widehat{X}}_{M}}=f_{1\times 1}^{Conv}\left( X \right) \\ 
 & \widehat{X}_{M-1}^{{{S}_{1}}}=\gamma \left( f_{3\times 3}^{DWConv}\left( {{\widehat{X}}_{M}} \right) \right) \\
 & \widehat{X}_{M-1}^{{{S}_{2}}}=\gamma \left( f_{5\times 5}^{DWConv}\left( {{\widehat{X}}_{M}} \right) \right) \\
 & \widehat{X}_{M}^{{{S}_{1}}}=\gamma \left( f_{3\times 3}^{DWConv}\left( Cat\left( \widehat{X}_{M-1}^{{{S}_{1}}},\widehat{X}_{M-1}^{{{S}_{2}}} \right) \right) \right) \\
 & \widehat{X}_{M}^{{{S}_{2}}}=\gamma \left( f_{5\times 5}^{DWConv}\left( Cat\left( \widehat{X}_{M-1}^{{{S}_{1}}},\widehat{X}_{M-1}^{{{S}_{2}}} \right) \right) \right)  \\ 
 & \widehat{X}=f_{1\times 1}^{Conv}\left( Cat\left( \widehat{X}_{M}^{{{S}_{1}}},\widehat{X}_{M}^{{{S}_{2}}} \right) \right)
\end{align}

where $\gamma \left( {} \right)$ denotes the ReLU activation function. After the multi-scale feature fusion module, the encoder and decoder features are fed into a masking module for spatial saliency modeling. This module applies three different channel-gating filters to identify the most discriminative regions in the spatial domain, and uses a Softmax activation function to implement a weighted strategy that emphasizes high-response areas. This saliency design effectively handles challenges such as blurred lesions and indistinct contours in medical images. Next, the filtered features are added together and further fused using dilated convolution with a dilation rate of 2. This expands the receptive field, improves the capture of lesion shapes and structural boundaries, and introduces richer contextual information without compressing the feature map resolution. Finally, the fused features are processed through a normalization layer and a pointwise convolution for channel alignment and stabilization of feature distributions, preventing gradient vanishing and improving training stability.

\begin{table}[t]
\centering
\small
\begin{tabular}{@{}l|llll@{}}
\toprule  \hline
                  Methods    & DSC   & RV    & Myo      & LV  \\  \hline \hline
                 Swin-UNet        & 88.07  & 85.77 & 84.42 & 94.03    \\
                 TransUnet        & 89.71  & 86.67 & 87.27 & 95.18    \\
                 Cascaded MERIT     & 91.85  & 90.23  & 89.53  & 95.80    \\
                 PVT-GCASCADE         & 91.95  & 90.31  & 89.63  & 95.91    \\
                 DMSA-UNet       & 92.28  & 90.32  & 90.49  & 96.02    \\
                 EMCAD     & 92.12  & 90.65  & 89.68  & 96.02    \\
                 AD-LA Former   & 90.09 & 88.68  & 88.94   & 95.30 \\
                 CSWin-UNet      &91.46  & 89.68  & 88.94 & 95.76 \\ \hline
 \textbf{Ours}            & {\textbf{92.75} } & {\textbf{91.18} } & {\textbf{90.40} } &{\textbf{96.67} } \\ \hline \hline 
\end{tabular}
\caption{Comparison experiments on the ACDC dataset.}
\label{tab:ACDC}
\end{table}

\section{Experiments}
\subsection{Experimental Setup}
We implemented our model using PyTorch 1.11.0 and conducted all experiments on a single NVIDIA A100 GPU with 40GB memory. Following the experimental settings of EMCAD, we adopted PVTv2-b2 pre-trained on ImageNet as the encoder. The learning rate and weight decay were set to 1e-4, the AdamW optimizer was used during training, normalized masks [0,1], and no augmentation. The batch size was fixed at 12. We trained the model for 200 epochs on the ISIC 2017 and ISIC 2018 datasets, and for 300 and 400 epochs on the Synapse and ACDC datasets, respectively.

\subsection{Datasets}
\subsubsection{Skin Lesion Segmentation}
We used the ISIC 2017 and ISIC 2018 datasets, released by the International Skin Imaging Collaboration. ISIC 2017 contains about 2,000 high-resolution dermoscopic images with precise lesion annotations, mainly for boundary segmentation evaluation. ISIC 2018 expands to 2,594 images with more lesion types, providing a benchmark for multi-class segmentation and generalization assessment.

\subsubsection{Synapse Multi-Organ Segmentation}
The Synapse dataset, from the MICCAI 2015 Multi-Atlas Labeling Beyond the Cranial Vault challenge, includes 30 abdominal CT scans with pixel-level annotations of eight organs, such as the liver, pancreas, kidneys, and spleen. Its standardized imaging and high-quality masks make it a widely used benchmark for multi-organ segmentation, cross-structure recognition, and generalization studies.

\begin{figure*}
    \centering
    \includegraphics[width=1\linewidth]{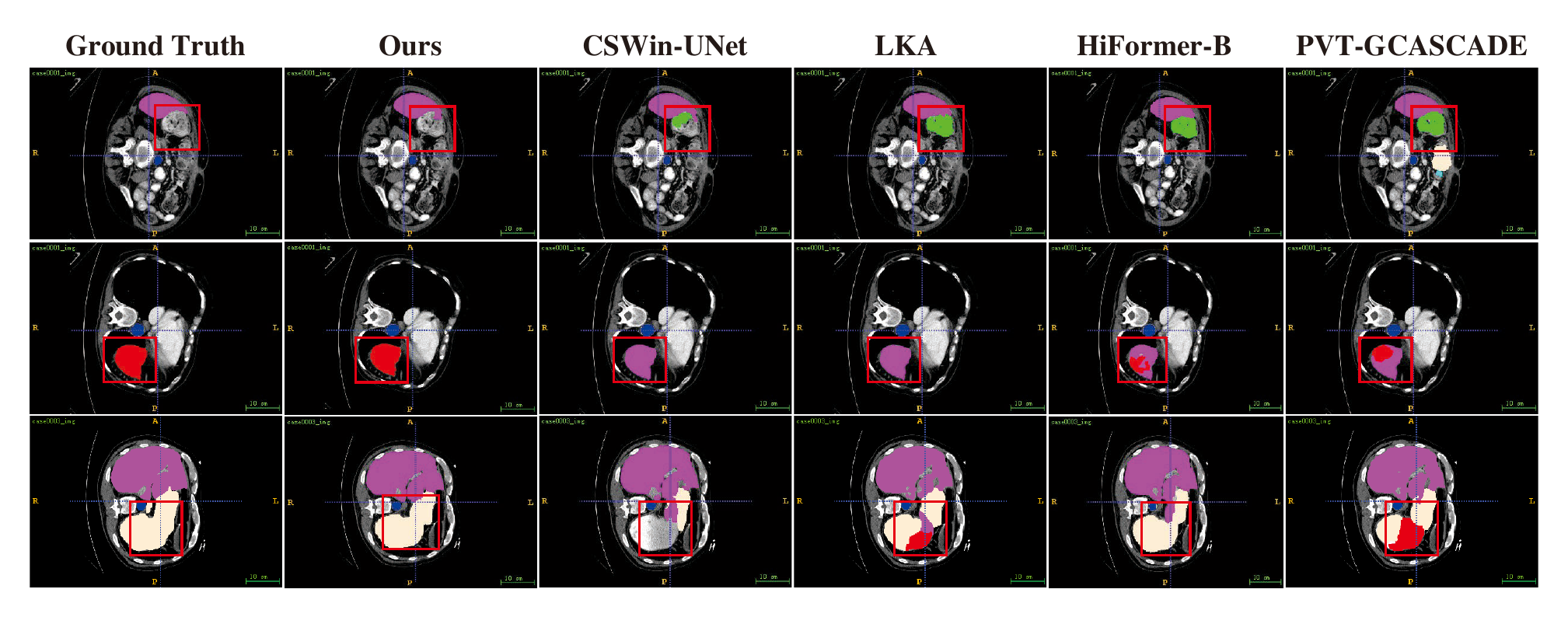}
    \caption{Comparison of the performance of our model with the visualization results of other models on the Synapse dataset.}
    \label{fig:SY}
\end{figure*}

\begin{figure}
    \centering
    \includegraphics[width=0.9\linewidth]{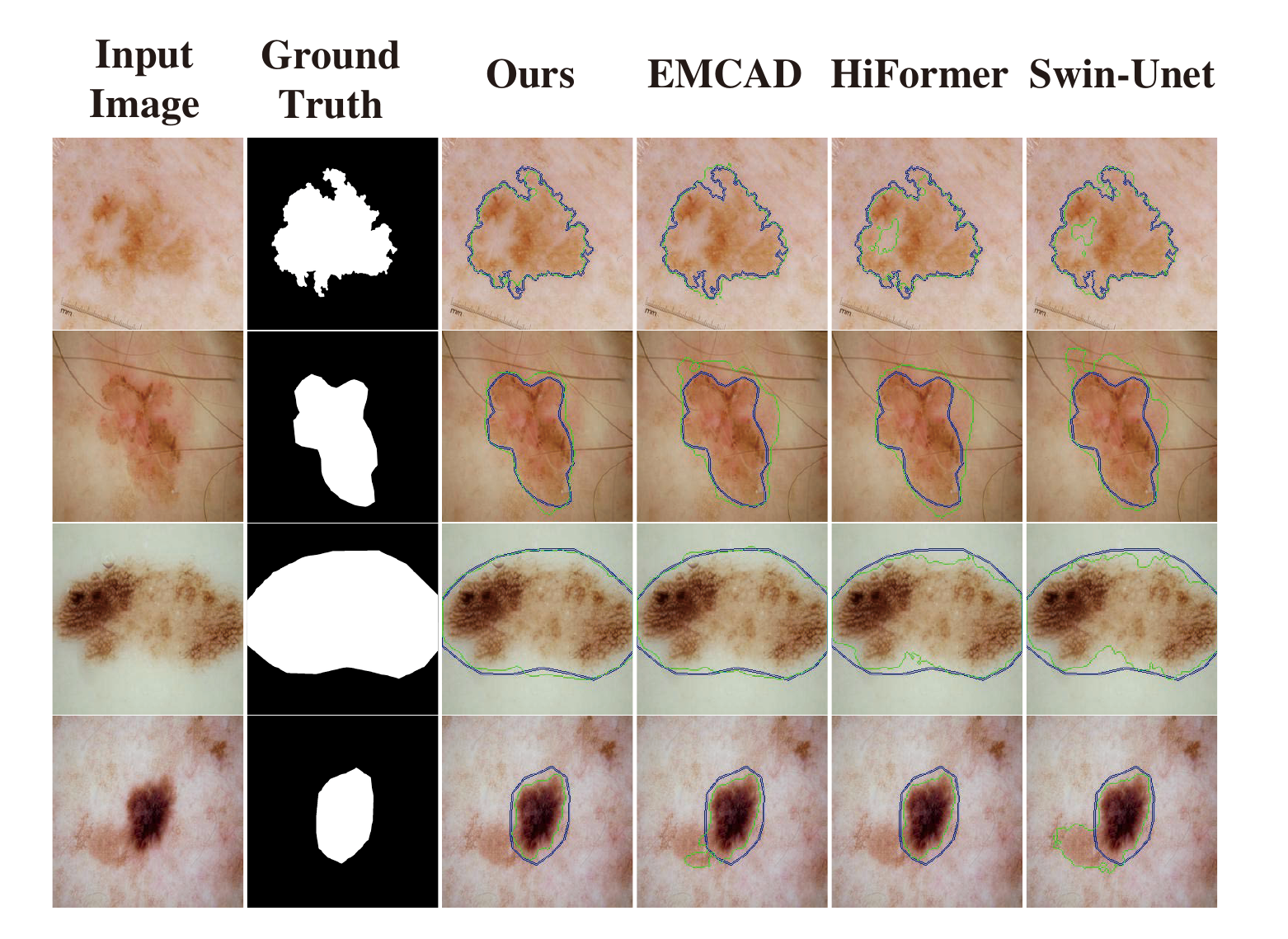}
    \caption{Comparison of segmentation results on the ISIC 2017 dataset with other previous methods.}
    \label{fig:ISIC2017}
\end{figure}

\subsubsection{ACDC Dataset}
The ACDC dataset, part of the MICCAI 2017 cardiac segmentation challenge, comprises cardiac MRI cine sequences from 100 subjects, including healthy and pathological cases (e.g., dilated and hypertrophic cardiomyopathy). It provides precise annotations for the left and right ventricles and myocardium, serving as a benchmark for cardiac structure segmentation and functional analysis.

\subsection{Quantitative and Qualitative Results}
On the Synapse dataset (Table \ref{tab:SY}), our model achieved an average DSC of 83.92\%, surpassing state-of-the-art methods such as EMCAD (83.63\%) and AD-LA Former (83.48\%), demonstrating a clear advantage in multi-organ segmentation. It achieved the best performance on key organs including the spleen (92.46\%), right kidney (86.47\%), and left kidney (89.26\%), highlighting its ability to capture complex organ boundaries and fine-grained structures. Although AD-LA Former obtained the highest score on the gallbladder (83.30\%), its overall average remains lower than ours. We additionally compared our model with other recent methods, including ScaleFormer \cite{scaleformer}, DAEFormer \cite{Dae-former}, PVT-CASCADE \cite{PVT-CASCADE}, LKA \cite{lka}, EGE-Unet \cite{Ege-unet}, UltraLightVM-UNet \cite{Ultralightvm-unet}, Cascaded MERIT \cite{CascadedMERIT}, VM-UNetV2 \cite{VM-UNET-V2}, and DMSA-UNet \cite{DMSA-UNet} across Synapse, ISIC 2017, ISIC 2018, and ACDC datasets. On ISIC 2017 (Table \ref{tab:17}), our model achieved the highest DSC (91.40\%) and ACC (97.26\%), outperforming EMCAD (90.06\%) and LKA (90.99\%), while maintaining a balanced sensitivity and specificity. On ACDC (Table \ref{tab:ACDC}), it achieved an average DSC of 92.75\%, with RV, Myo, and LV scores of 91.18\%, 90.40\%, and 96.67\%, exceeding DMSA-UNet (92.28\%) and Cascaded MERIT (91.85\%). On ISIC 2018 (Table \ref{tab:18}), it reached the best DSC (90.71\%) and ACC (96.62\%) and a competitive SE (93.34\%), second only to DMSA-UNet (97.14\%). These comprehensive results, further supported by detailed visualization comparisons (Figures \ref{fig:SY} and \ref{fig:ISIC2017}), confirm the robustness, structural awareness, and strong cross-task generalization of our model across multi-organ, cardiac, and skin lesion segmentation tasks, underscoring its effectiveness in handling diverse and challenging medical imaging scenarios.

\begin{figure}
    \centering
    \includegraphics[width=1\linewidth]{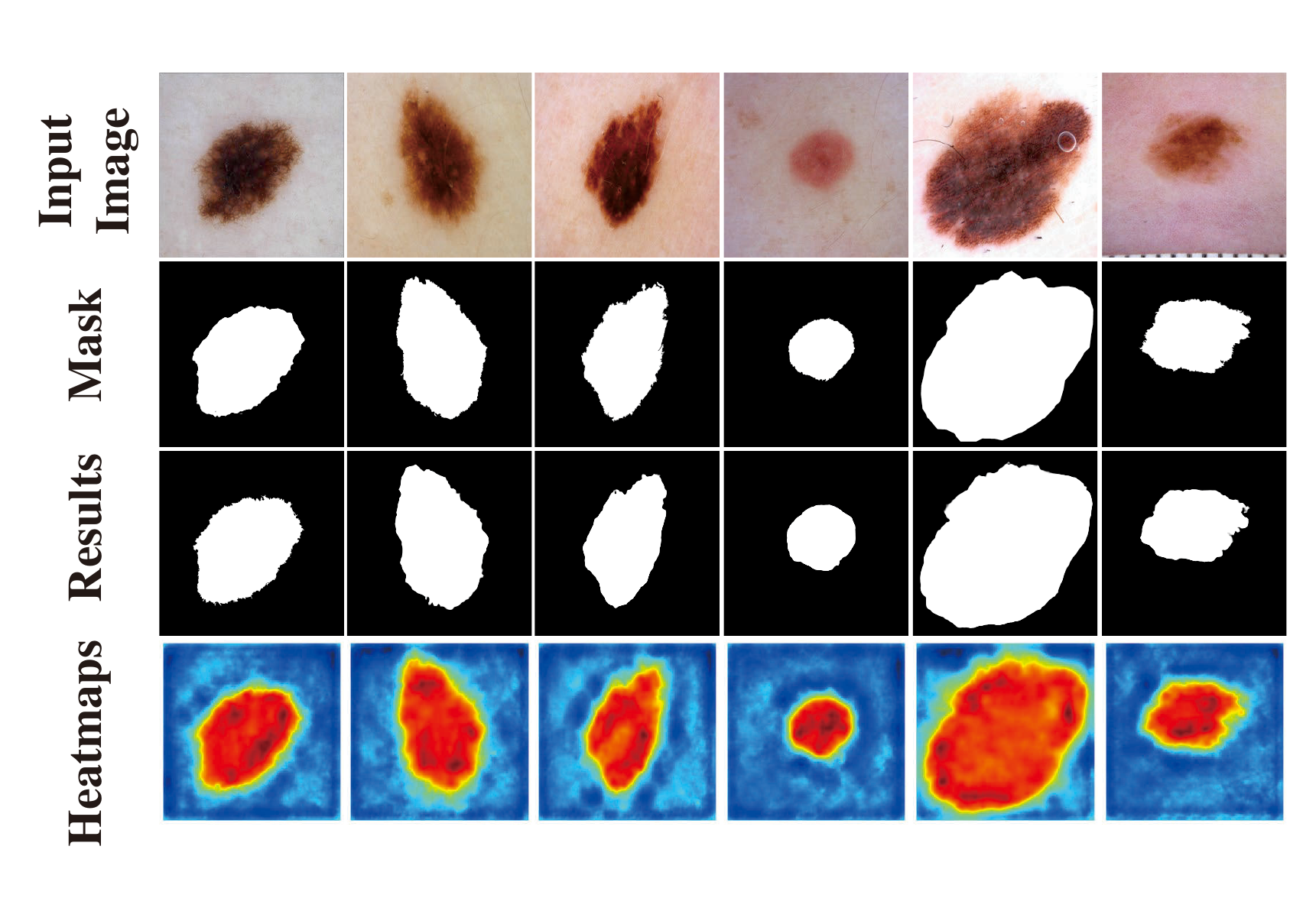}
    \caption{Attention heatmap visualization for the proposed model on ISIC 2018.}
    \label{fig:heatmaps}
\end{figure}

\begin{table*}[ht]
\centering
\begin{tabular}{@{}l|ll|llllllll@{}}
\toprule
\hline
\multirow{2}{*}{Methods} & \multicolumn{2}{c|}{Average}  & \multirow{2}{*}{Aor} & \multirow{2}{*}{Gal} & \multirow{2}{*}{LKid} & \multirow{2}{*}{RKid} & \multirow{2}{*}{Liv} & \multirow{2}{*}{Pan} & \multirow{2}{*}{Spl} & \multirow{2}{*}{Sto} \\ \cline{2-3}
                       & DSC ↑       & HD95  ↓                  
                       &                      &                       &                      
                       &                      &                      &                      
                       &                      &                     \\  \hline \hline
Baseline               & 81.35 & 20.42 & 86.78 & 64.11 & 87.48 & 83.96 & 93.82 & 62.00 & 89.88 & 82.77 \\
Baseline+ACFA           & 82.04 & 21.46 & 86.54 & 65.73 & 87.69 & 84.27 & 93.38 & 65.17 & 90.12 & 83.42 \\
Baseline+ACFA+TFFA      & 83.23 & 16.72 & 86.05 & 64.43 & 87.89 & 86.86 & 94.58 & 70.99 & 91.07 & 83.97 \\
Baseline+ACFA+TFFA+SMMM & 83.92 & 18.91 & 87.63 & 67.51 & 89.26 & 86.47 & 94.72 & 69.95 & 92.46 & 83.33 \\ \hline \hline
\end{tabular}
\caption{Ablation study on the impact of different modules on the model on the Synapse dataset.}
\label{tab:DSY}
\end{table*}

\begin{table}[ht]
\centering
\small
\begin{tabular}{@{}l|llll@{}}
\toprule
\hline
Method                  & DSC   & SE    & SP    & ACC    \\ \hline \hline
Baseline                & 85.95 & 83.68 & 98.96 & 96.05 \\
Baseline+ACFA           & 87.82 & 85.12 & 98.16 & 95.92 \\
Baseline+ACFA+TFFA      & 89.15 & 89.83 & 97.29 & 96.85 \\
Baseline+ACFA+TFFA+SMMM & 91.40 & 92.75 & 97.78 & 97.26  \\  \hline \hline
\end{tabular}
\caption{Study on the ablation of the impact of different modules on the ISIC 2017 dataset.}
\label{tab:ab17}
\end{table}

\begin{table}[ht]
\centering
\small
\begin{tabular}{@{}l|ll|llll@{}}
\hline \hline
\multirow{2}{*}{Fourier} & \multicolumn{2}{|c|}{Wavelet} & \multirow{2}{*}{DSC}   & \multirow{2}{*}{SE}    & \multirow{2}{*}{SP}    & \multirow{2}{*}{ACC}   \\ \cline{2-3}
        & Mexican Hat      & DoG      &       &       &       &       \\  \hline \hline
NO      & NO               & NO       & 90.32 & 89.31 & 97.73 & 95.37 \\
YES     & NO               & NO       & 90.48 & 91.43 & 97.27 & 95.74 \\
YES     & YES              & NO       & 90.57 & 92.63 & 96.41 & 96.15 \\
YES     & YES              & YES      & 90.71 & 93.34 & 96.18 & 96.62  \\  \hline \hline
\end{tabular}
\caption{Ablation study on the effects of different components within the TFFA module.}
\label{tab:TFFA}
\end{table}

\begin{table}[ht]
\centering
\small
\begin{tabular}{@{}lll|llll@{}}
\toprule
\hline
ACFA & TFFA & TFFA & Params (M) & Complexity (GMac) \\   \hline \hline
NO   & NO   & NO   & 25.07      & 11.85                           \\
YES  & NO   & NO   & 30.67      & 13.35                           \\
YES  & YES  & NO   & 32.01      & 13.75                           \\
YES  & YES  & YES  & 42.52      & 18.29                           \\  \hline \hline
\end{tabular}
\caption{Computational cost and complexity of modules on synapse dataset.}
\label{tab:pc}
\end{table}

\section{Ablation Studies}
To validate the effectiveness of the proposed decoder, we conducted ablation experiments on the Synapse and ISIC 2017 datasets, as shown in Tables \ref{tab:DSY}, \ref{tab:ab17}, \ref{tab:TFFA}and \ref{tab:pc}.

\subsection{Effect of Different Components on Synapse}
Results on the Synapse dataset confirm that each component contributes to performance improvement. Introducing ACFA increased parameters by only 5.6M and computation by 1.5 GMac, while boosting DSC to 82.04\%. This demonstrates that directional awareness and channel–spatial fusion strengthen structural perception. Incorporating TFFA further improved DSC to 83.2\% and reduced HD95 to 16.72, indicating that joint spatial–frequency modeling complements global semantics and local boundary learning. When all modules were integrated, the decoder achieved 83.92\% DSC with 42.52M parameters and 18.29 GMac, delivering the best segmentation of complex organs such as the liver and kidneys while maintaining efficiency.

\subsection{Effect of Different Components on ISIC 2017}
We also conducted ablation studies on the ISIC 2017 dataset. The baseline achieved 85.95\% DSC. Adding ACFA raised it to 87.82\%, highlighting enhanced directional and regional discrimination. When TFFA was incorporated, DSC reached 89.15\% and SE increased to 89.83\%, verifying that spatial–frequency fusion improves edge and texture representation. Integrating SMMM yielded the best performance (DSC 91.4\%, SE 92.75\%, ACC 97.26\%), demonstrating that joint multi-module learning enhances lesion delineation and boundary accuracy.

\subsection{Attention Heatmap Analysis on the ISIC 2018 Dataset}
Attention heatmaps on the ISIC 2018 dataset (Fig. \ref{fig:heatmaps}) show that the proposed decoder captures lesion boundaries, textures, and internal structures effectively. ACFA strengthens directional edge awareness, TFFA balances global and local cues through Fourier–wavelet fusion, and SMMM alleviates spatial detail loss during feature aggregation. Their synergy enables precise focus on lesion regions, smooth boundary prediction, and improved generalization.

\subsection{Internal Ablation Study of the TFFA Module}
We conducted an internal analysis of the TFFA module and validated the contribution of each frequency component. Fourier modeling enhances global structural representation, while adding the Mexican Hat wavelet improves local edge sensitivity. Combining both DoG and Mexican Hat wavelets achieved the best results (DSC 90.71\%, SE 93.34\%), confirming that multi-scale spatial–frequency fusion effectively refines boundary perception. 

\section{Conclusion}
We propose a novel decoder for medical image segmentation that addresses challenges in edge detail modeling, long-range dependency capture, and multi-scale feature fusion. The framework integrates ACFA for directional and structural awareness, TFFA for joint Wavelet–Fourier–Spatial feature modeling, and SMMM for multi-scale skip fusion with saliency masking. Overall, the decoder achieves significant segmentation accuracy gains through joint directional–frequency–structural modeling, providing an effective and practical solution for high-precision medical image segmentation.

\section*{Ethical Statement}
All datasets used in this study are publicly available and de-identified. Our experiments comply with the terms of use provided by the data providers, ensuring no personally identifiable information is disclosed. This research adheres to standard ethical guidelines for the use of medical imaging data.

\section*{Acknowledgements}
This work was supported in part by the following: the National Natural Science Foundation of China under Grant Nos. U24A20219, 62272281, U24A20328, U22A2033, 62576193, the Special Funds for Taishan Scholars Project under Grant Nos. tsqn202306274, tsqn202507240, the Yantai Natural Science Foundation under Grant No. 2024JCYJ034, the Youth Innovation Technology Project of Higher School in Shandong Province under Grant No. 2023KJ212, and the Natural Science Foundation of Shandong Province under grant No. ZR20250C712.

\bibliography{main}


\end{document}